\documentclass[10pt,twocolumn,letterpaper]{article}

\usepackage{cvpr}
\usepackage{times}
\usepackage{epsfig}
\usepackage{graphicx}
\usepackage{amsmath}
\usepackage{amssymb}

\usepackage{mathtools}
\usepackage{multirow}
\usepackage{diagbox}
\usepackage{booktabs}
\usepackage{bm}

\usepackage[breaklinks=true,bookmarks=false]{hyperref}

\cvprfinalcopy 

\def\cvprPaperID{6624} 

\begin{document}

\title{Iteratively-Refined Interactive 3D Medical Image Segmentation\\ with Multi-Agent Reinforcement Learning}

\author{Xuan Liao\textsuperscript{1}  
,
Wenhao Li\textsuperscript{2}\thanks{Equal contribution}
,
Qisen Xu\textsuperscript{2}\footnotemark[1]
,
Xiangfeng Wang\textsuperscript{2}
,
\\Bo Jin\textsuperscript{2}
,
Xiaoyun Zhang\textsuperscript{1}
,
Ya Zhang\textsuperscript{1}
,
and Yanfeng Wang\textsuperscript{1}
\\\textsuperscript{1}Shanghai Jiao Tong University
,
\textsuperscript{2}East China Normal University
}

\maketitle

\begin{abstract}
Existing automatic 3D image segmentation methods usually fail to meet the clinic use. 
Many studies have explored an interactive strategy to improve the image segmentation performance by iteratively incorporating user hints. 
However, the dynamic process for successive interactions is largely ignored. 
We here propose to model the dynamic process of iterative interactive image segmentation as a Markov decision process (MDP) and solve it with reinforcement learning (RL). 
Unfortunately, it is intractable to use single-agent RL for voxel-wise prediction due to the large exploration space. 
To reduce the exploration space to a tractable size, we treat each voxel as an agent with a shared voxel-level behavior strategy so that it can be solved with multi-agent reinforcement learning. 
An additional advantage of this multi-agent model is to capture the dependency among voxels for segmentation task. 
Meanwhile, to enrich the information of previous segmentations, we reserve the prediction uncertainty in the state space of MDP and derive an adjustment action space leading to a more precise and finer segmentation.
In addition, to improve the efficiency of exploration, we design a relative cross-entropy gain-based reward to update the policy in a constrained direction.
Experimental results on various medical datasets have shown that our method significantly outperforms existing state-of-the-art methods, with the advantage of fewer interactions and a faster convergence.
\end{abstract}

\section{Introduction}
Medical image segmentation has been widely recognized as an essential procedure for subsequent medical image processes such as structural and functional analysis, diagnosis and treatment. The traditional dense manual annotation is extremely inefficient for 3D medical images and its performance highly depends on the physician's experience. With the development of convolutional neural networks (CNNs), automatic segmentation has greatly improved the efficiency of medical image process \cite{milletari2016v,ronneberger2015u,zhou2018unet++}. However, the accuracy and robustness for the current automatic methods still need to be improved for practical clinic use.

To get a better segmentation, interactive image segmentation \cite{bredell2018iterative,cciccek20163d,wang2018deepigeos} is introduced to integrate user hints (mostly in the form of points, scribbles and bounding boxes). This kind of interactive methods has become a popular research direction because it improves the performance of segmentation by adding new labeling constraints to the prediction model. Normally, a one-time interaction might not ensure the segmentation accuracy. Therefore, many existing methods are compatible with the iteratively-refined mode: the operator provides new hints according to the current result to refine the segmentation until it is satisfactory. Moreover, to reduce the number of interactions, the existing works introduce the idea of replacing the initial hints with an automatically-obtained coarse segmentation \cite{bredell2018iterative,wang2018deepigeos}. Note that in this paper, we refer to such methods incorporating a coarse segmentation in the initial input as \textit{\textbf{update methods}} and we will focus on this kind of methods in this paper.

Concerning the current update methods, there exist two main issues: 1) They usually ignore the dynamic process for successive interactions. Although the segmentation can be iteratively refined, the model always treat the segmentation of each refinement step in isolation, with an absence of the previous information.
2) Another problem is the loss of prediction uncertainty when using the binary segmentation result, instead of a segmentation probability for each voxel, as part of the model input. The rounding from dense segmentation probability to binary segmentation prediction may cause quantization error and accuracy loss.

To tackle the above two issues, this paper proposes a novel interactive medical image segmentation update method called \textit{Iteratively-Refined interactive 3D medical image segmentation via Multi-agent Reinforcement Learning (IteR-MRL)}. 
We formulate the dynamic process of iterative interactive image segmentation as an MDP.
Specifically, at each refinement step, the model needs to decide the labels of all voxels, according to the previous segmentations and supervision information from the interaction.
After that, the model will get the feedback according to pre-defined measurement of segmentation, and the above process will be repeated until the maximum number of interactions is reached.
We then adopt the RL methods to solve above MDP, that is, to find the segmentation strategy to maximize the accumulated feedbacks received at each refinement step.
However, it will be intractable to use single-agent RL for voxel-wise prediction due to the large exploration space.
In addition, considering that the voxels in the segmentation task are interdependent, they can achieve better segmentation by a more comprehensive grasp of the surrounding information.
To reduce the exploration space to a tractable size and explicitly model the dependencies between voxels, we introduce the multi-agent reinforcement learning (MARL) method.
We treat each voxel as an agent which decides its own label. 
All agents share the same policy and collaborate with each other through convolutional kernels.
Meanwhile, instead of considering the difference between the current prediction and the ground truth, we design a relative cross-entropy gain-based reward to prompt agents to explore more efficiently.
Specifically, the algorithm gives a positive reward for an improvement and vice versa at each refinement step, so that the new prediction can be forced to outperform the previous one.
Compared with supervised methods, such RL-based methods have the advantage of a faster refinement convergence.
The problem of prediction uncertainty loss in existing works caused by segmentation map binarization can be settled by adopting segmentation probability rather than binary segmentation as part of the RL state.
This augmented state space also derives an adjustment action space leading to a more precise and finer segmentation.
Then the segmentation refinement procedure can be regarded as a series of actions to adjust the segmentation probability with a certain level. 
In this way, the prediction uncertainty is reserved and the algorithm explores in a finer granularity and a denser space.

The experimental results indicate that the proposed IteR-MRL is robust to different initial segmentations and various medical datasets. Given the same initial segmentations, our proposed interactive algorithm surpasses the state-of-the-art update methods on different 3D medical image segmentation datasets including the images of brain tumor, heart and prostate. We summarize our contributions as follows:
\begin{itemize}
    \item We formulate the interactive image segmentation task as an MDP and propose a novel voxel-wise interactive segmentation framework based on MARL for 3D medical images, enabling more effective utilization of user interaction.

    \item We propose to reserve the prediction uncertainty via the segmentation probability, which can enrich the information of previous segmentations and lead to a more precise and finer adjustment.

    \item Extensive experiments show that the segmentation is significantly improved over the iteration sequence with only a few interactions and a rapid convergence, by considering the relative gain between two successive steps.
\end{itemize}

\section{Related work}
Interactive image segmentation has been widely applied to both natural \cite{boykov2001interactive,xu2016deep} and medical images \cite{rajchl2016deepcut,wang2018interactive,wang2018deepigeos}. ``Interactive'' refers that the operator provides some hints to the segmentation model to achieve a better result. This section will briefly review the existing works.

\subsection{Graph-based interactive image segmentation}
Traditional methods make use of low-level features such as the histogram and similarities between pixels. GraphCut \cite{boykov2001interactive} and GrabCut \cite{rother2004grabcut} incorporate user hints into Max-Flow Min-Cut algorithm\cite{boykov2004experimental}. 
DenseCRF \cite{krahenbuhl2011efficient} considers pixel relations from neighbors to all pixel pairs. \cite{criminisi2008geos} proposes to use geodesic distance to calculate the distance between pixels, which is sensitive to contrast and suitable for medical images. \cite{wang2016slic} introduces a segmentation method for fetal MRI by learning from user annotations in only one slice.

\subsection{CNN-based interactive image segmentation}
Recently, using convolutional neural networks (CNNs) has become popular for computer vision problems. Many CNN-based methods have developed for interactive image segmentation tasks. \cite{xu2016deep} is the first one to use CNN in interactive image segmentation. \cite{rajchl2016deepcut} replaces the Gaussian mixture model (GMM) in GrabCut with a CNN for MRI segmentation. Another work, 3D U-Net \cite{cciccek20163d} learns to produce a complete segmentation from sparsely-annotated slices of one 3D medical image. In order to save the budget of the initial user hints, the following methods, known as update methods, choose to take an automatically-produced segmentation as part of the model initial input. \cite{wang2018deepigeos} proposes a two-stage method called DeepIGeoS to refine the segmentation using the initial coarse segmentation in input. However, the refined segmentations after the first step cannot be efficiently used in this model.
\cite{bredell2018iterative} extends DeepIGeoS to an iterative version: Inter-CNN, which iteratively refines the previous refined binary prediction in both training and testing stages. One of their problems is the ignorance of the dynamic process for successive interactions. 
Another problem is the accuracy loss caused by the quantization from probability to binary segmentation. 

\subsection{RL-based interactive image segmentation}
There are also some methods using RL to explicitly model the dynamic process in interactive image segmentation tasks. 
SeedNet \cite{song2018seednet} uses an RL agent to simulate the user behavior which gives the hints to the segmentation model. Since our method uses RL as the segmentation model to predict the segmentation, our method is orthogonal with it. Polygon-RNN \cite{castrejon2017annotating} identifies the object segmentation as a polygon. 
Their model produces vertices sequentially until the polygon is closed. 
The user can contribute by adjusting the vertices. 
Based on this work, Polygon-RNN++ \cite{acuna2018efficient} develops a faster and more accurate algorithm by combining RL with graph neural network. 
However, these polygon-based methods cannot be applied to our tasks because of the incompatibility of the 3D images with polygon segmentation, and the extreme large action space even with the meshing strategy.

\section{Methodology}

In this section, we formulate the interactive image segmentation as a MDP and propose a novel MARL-based interactive medical image segmentation method to exploit the interaction information more efficiently.

\subsection{Overview}
\begin{figure}
\centering
\includegraphics[width=3.2in]{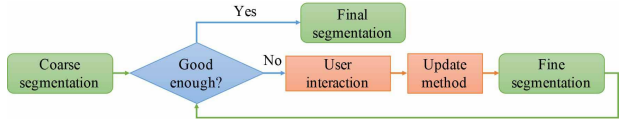}
\caption{The flow chart of iteratively-refined interactive image segmentation approach. Given a coarse segmentation, the method iteratively refines it with user interaction until the fine segmentation is good enough.}
\label{overview}
\end{figure}

In our work, we propose an iteratively-refined framework based on update methods, as shown in Fig. \ref{overview}, which iteratively refines a coarse initial segmentation by integrating user interactions in order to get a more precise segmentation result. The initial segmentation can be obtained from any accessible segmentation methods.

As discussed in Sections 1 and 2, the main issue of the existing supervised learning-based algorithms is that they split the whole image refinement process into isolated steps. To address this problem, we adopt RL to explicitly capture the relation between successive predictions by designing the reward as the relative improvement. 
As the large state space and action space of voxel-wise prediction and the necessity of the collaborations between interdependent voxels, we use the idea of MARL: each voxel in a 3D image is regarded as an agent. 
The work PixelRL \cite{furuta2019fully} also sees each pixel as an agent, but it focuses on general image processing tasks without human interaction. 
In contrast, the interactive image segmentation task is more suitable to adopt RL due to its intrinct sequentiality.
Unlike dealing with non-interactive image processing tasks, we aim to better consider and effectively utilize external supervision signal from the user during the interaction.

\begin{figure*}
\centering
\includegraphics[width=0.9\textwidth]{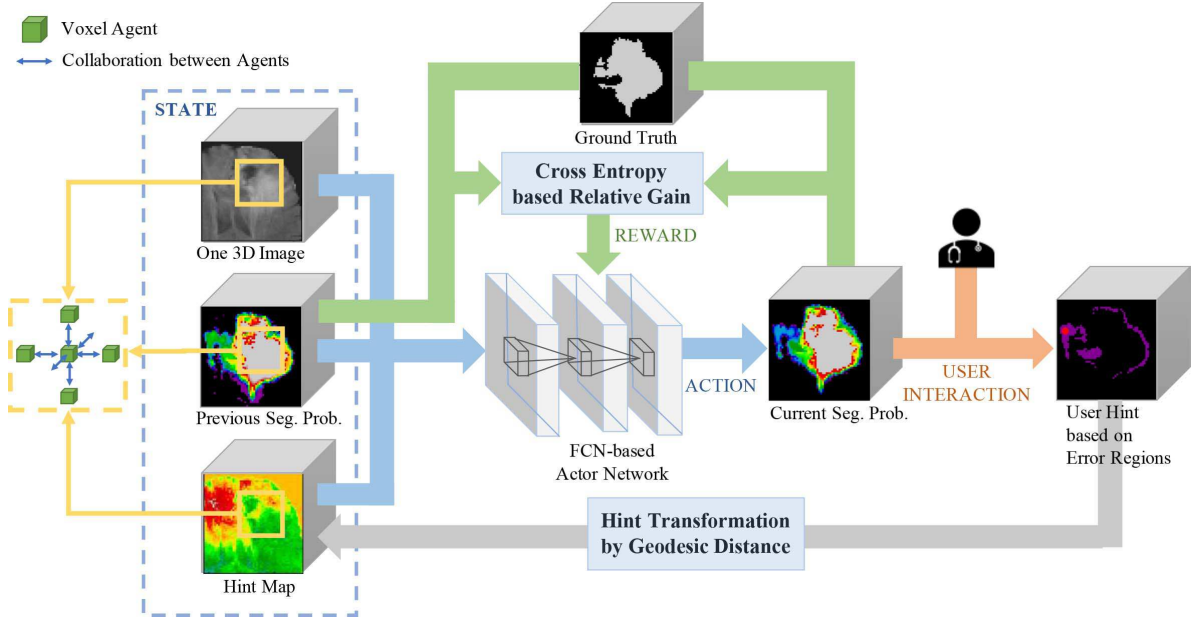}
\caption{Overview of Iteratively-Refined interactive 3D medical image segmentation algorithm based on MARL (IteR-MRL). At each refinement step, the state containing image, previous segmentation probability and the hint map is feeded into the actor network, then the actor network produces current segmentation probability derived by its output actions. Next, the user gives back hint clicks (the red point) based on error regions and new hint map is generated by hint transformation. At every step, the reward is determined by the relative gain between previous and current segmentation cross entropy. In the setting of our method, voxels are regarded as agents who collaborate with each other.}
\label{framework}
\end{figure*}

Fig. \ref{framework} introduces the framework of our proposed method IteR-MRL. By utilizing the original 3D image, the previous segmentation probability and the interactive information as the state, the actor network in the middle gives an update to the segmentation probability and produces a new one. 
Note that the previous segmentation probability is come from the previous update iteration and the interactive information is the hint map transformed from user hints which will be introduced in Subsection 3.2.
The actor network outputs actions of agents which adjust previous segmentation probability and generate current segmentation probability.
Afterwards, there are two subsequent operations for the current segmentation probability. 
On the one side, it gives back a reward signal to the network, by computing the relative gain between previous and current cross entropy based on ground truth and successive segmentation probabilities, for parameter updating. 
On the other side, it is presented to the user and the user provides feedback, \ie clicks on object or background for wrongly predicted areas.
The click is represented as a red point on Fig. \ref{framework}, which is enlarged for visualization. 
Generally, with a coarse segmentation probability produced by the initial method (any segmentation method) as its initial segmentation, IteR-MRL iteratively refines the segmentation probability until the segmentation is satisfactory. 
In addition, the actor network employed here is designed for MARL and it regards voxels on the 3D image as agents who collaborates with each other.

It should be noticed that instead of quantizing the segmentation probability to binary segmentation prediction like previous methods \cite{bredell2018iterative,wang2018deepigeos}, here we directly use the segmentation probability as the previous segmentation information and feed it into the model. 
The segmentation probability is introduced to enrich the previous segmentation information and achieve more accurate results.
With the segmentation probability, we can derive an adjustment action space leads to a more precise and finer segmentation comparing with the binary segmentation quantization.
Specifically, we can adjust the segmentation probability at each step and choose the best adjustment magnitude from a set of various scales. 
The adjustment action for one agent in MARL model is based on both its own and the neighbors' states.

\subsection{Multi-agent RL framework for interactive image segmentation}

In this subsection, we describe the MARL setting for interactive image segmentation. Let $\bm{x} = (x_1, \cdots, x_N)$ be one arbitrary image in the dataset and $x_{i}$ is the $i_{th}$ voxel of $\bm{x}$. We treat each $x_{i}$ as an agent whose policy is defined as $\pi_{i} (a_i^{(t)}|s_i^{(t)})$. $s_i^{(t)}$ and $a_i^{(t)}$ are the state (image, previous segmentation probability, user interaction) and action (adjustment to previous probability) for $x_{i}$ at the step $t$; $a_i^{(t)} \in \mathcal{A}$ and $\mathcal{A}$ is the action set; $s_i^{(t)} \in \mathcal{S}$ and $\mathcal{S}$ is the state set. By using convolutional kernels, one agent can access to its neighbors' states as well, where neighbors are considered as near voxels.

From the point of view of the whole image, the previous segmentation is refined to a new one. By taking the global action $\bm{a}^{(t)} = (a_1^{(t)}, \cdots, a_N^{(t)})$, the image agent transfers to the global state $\bm{s}^{(t+1)} = (s_1^{(t+1)}, \cdots, s_N^{(t+1)})$ and gets the global reward $\bm{r}^{(t)} = (r_1^{(t)}, \cdots, r_N^{(t)})$.

We now define the state, action and reward of a single agent $x_{i}$ in IteR-MRL.\\

\noindent \textbf{State.}
For our problem formulation, the state for voxel agent $x_{i}$ at the step $t$ is the concatenation of its voxel value $b_{i}$, its previous segmentation probability $p_{i}^{(t)}$ to be object label and its two values on hint maps $h_{+,i}^{(t)}$ and $ h_{-,i}^{(t)}$: $s_{i}^{(t)}=[b_{i}, p_{i}^{(t)}, h_{+,i}^{(t)}, h_{-,i}^{(t)}]$ with $p_{i}^{(t)} \in [0,1]$. For the initial state $s_{i}^{(0)}$, the initial coarse segmentation probability denotes initial probability $p_{i}^{(0)}$.

Now we discuss the generation of a whole hint map. Concerning the user interaction at step $t$, the hint map $\bm{h}^{(t)}$ is transformed from the user's hints which are in the form of click points. By giving a hint point through a single click, the user indicates that the area around it is one error region. Intuitively, the closer one point is to the hint point, the more likely its label is mispredicted. Hence, the hint map is introduced to show the radiation area of the hint and spread the local interaction to the whole image. The number and positions of hints are chosen according to the user interaction rule. Actually, there are two channels of hint map both with the same size as the image: the object hint map $\bm{h}_{+}^{(t)}$ and the background hint map $\bm{h}_{-}^{(t)}$, respectively generated from the object hint set $hs_{+}^{(t)}$ (hints on object) and the background hint set $hs_{-}^{(t)}$ (hints on background). Hence, the user hint map is the concatenation of these two hint maps: $\bm{h}^{(t)}=[\bm{h}_{+}^{(t)}, \bm{h}_{-}^{(t)}]$. For one hint map $\bm{h}_{\ell}^{(t)}, \ \ell \in \{+, -\}$, we define that $\bm{h}_{\ell}^{(t)} = (h_{\ell,1}^{(t)}, \cdots, h_{\ell,N}^{(t)})$. The element $h_{\ell,i}^{(t)}$ on the hint map $\bm{h}_{\ell}^{(t)}$ is calculated by the minimum distance between $x_i$ and the corresponding hint set $hs_{\ell}^{(t)}$:
\begin{equation}
h_{\ell,i}^{(t)} \ = \ \min_{\forall x_j \in hs_{\ell}^{(t)}} \mathcal{M}(x_i, x_j),\\\ \ell \in \{+, -\},
\end{equation}
where $\mathcal{M}$ is a function to measure the distance between two voxels. Previous related works adopt several distance-measuring methods including geodesic \cite{criminisi2008geos}, Gaussian and Euclidean distance. In the paper, we use the geodesic distance-based hint map to measure distances. The distance between two voxels is the minimum value of the sum of color gradients across all the paths connecting these two voxels. (See the hint map in Fig. \ref{framework}.)\\

\noindent \textbf{Action.}
While previous works \cite{bredell2018iterative,wang2018deepigeos} output directly the segmentation probability from the network, we here predict the adjustment amount based on the previous probability as actions to make the result more stable without abrupt changes. The action $a_{i}^{(t)} \in \mathcal{A}$ for $x_{i}$ at time step $t$ is to adjust the previous segmentation probability $p_{i}^{(t)}$ by a certain amount $a_{i}^{(t)}$. Hence, the segmentation probability $p_{i}^{(t+1)}$ after taking the action $a_{i}^{(t)}$ is:
\begin{align} 
p_{i}^{(t+1)} \ &=\ \mathcal{C}_0^1(p_{i}^{(t)} + a_{i}^{(t)}),\\
\label{clip}
\mathcal{C}_a^b(x) \ &=\ \min(\max(x, a), b),
\end{align}
where $\mathcal{C}_a^b(x)$ is a function to clip the value of $x$ from $a$ to $b$. $p_{i}^{(t+1)}$ is constrained to $[0,1]$ for it represents a probability. The action set $\mathcal{A} = \{\mathcal{A}_k\} \ (k = 1,2, \cdots, K)$ contains $K$ actions, allowing the agent to adjust the probability to various degrees under different situations. For example, it is reasonable to make a larger  adjustment to a voxel when it is closer to a hint click. Additionally, one voxel tends to take one certain adjustment action when most of its neighbor voxels choose this action.\\

\noindent \textbf{Reward.}
To improve the efficiency of exploration, we design a relative cross-entropy gain-based reward to update the model in a constrained direction. 
Specifically, the reward is designed as the relative improvement from the previous segmentation to the current one, which is the decreased amount of the cross entropy $\mathcal{X}_{i}$ between the ground truth $y_{i}$ and the segmentation probability $p_{i}$:
\begin{equation} 
\label{reward}
r_{i}^{(t)} \ =\ \mathcal{X}_{i}^{(t-1)} - \mathcal{X}_{i}^{(t)},
\end{equation}
where 
\begin{equation} 
\label{cross entropy}
\mathcal{X}_{i}^{(t)} \ =\ -y_{i}\log(p_{i}^{(t)})-(1-y_{i})\log(1-p_{i}^{(t)}).
\end{equation}

With \eqref{reward}, the agent gets a positive reward in the case its probability moves closer towards the true voxel label and vice versa. Instead of a distant goal, the relative gain provides the agent with a baseline to compare and surpass.

In general, the accumulated reward of one interactive sequence is
\begin{equation}
R_i \ = \ \textstyle\sum_{t=1}^{T}\gamma^{t-1} r_{i}^{(t)},
\end{equation}
where $T$ is the total step number and the discount factor $\gamma$ takes a value in $(0, 1]$.

\subsection{Network and training}
\begin{figure}
\centering
\includegraphics[width=3in]{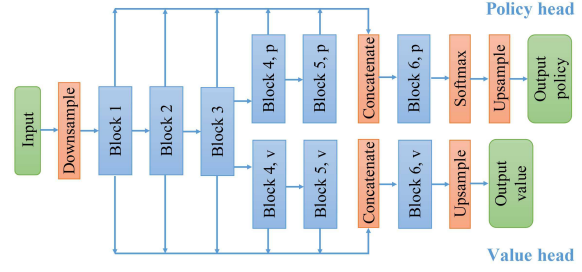}
\caption{The network architecture for IteR-MRL. The policy and value heads share the low-level features and extract their own high-level features.}
\label{architecture}
\end{figure}

For fair comparison, the interactive network architecture of \cite{wang2018deepigeos} named R-net is adopted as the backbone to our algorithm and all other baseline methods. We adapt the network to the one in Fig. \ref{architecture} in order to fit the RL training algorithm: asynchronous advantage actor-critic (A3C) \cite{mnih2016asynchronous}. The network firstly uses three 3D convolutional blocks to extract low-level features. Then, the network is divided into two heads: policy head and value head. Both of the heads have three 3D convolutional blocks to extract specific high-level features. The functionality of the policy head is to predict the distribution of action probabilities under a known state. In our case, given the image, hint maps and previous segmentation probability, the policy head predicts how likely it is to take each scale of adjustment to the previous segmentation probability. The functionality of the value head is to estimate the value of the current state. Specifically, the value head evaluates how good the current combination of the image, hint maps and the previous segmentation probability is.

We respectively use $\theta_p$ and $\theta_v$ to denote the parameters of the policy and value heads. The input of the network is the state at time step t: $\bm{s}^{(t)}$. The value head outputs the estimated value of the current state $V(\bm{s}^{(t)})$. The gradient for $\theta_v$ is computed by:
\begin{align} 
d\theta_v \ &= \ \nabla_{\theta_v}A(\bm{s}^{(t)}, \bm{a}^{(t)})^2,\\
A(\bm{s}^{(t)}, \bm{a}^{(t)}) \ &= \  \sum_{k=t}^{T}\gamma^{k-t} \overline{r}^{(k)} - V(\bm{s}^{(t)}),
\end{align}
where $\overline{r}^{(k)}$ is the mean reward of all voxels at time step $k$. $A(\bm{s}^{(t)}, \bm{a}^{(t)})$ is the advantage at time step $t$ of taking $\bm{a}^{(t)}$ in condition of state $\bm{s}^{(t)}$, which indicates the actual accumulated reward without being affected by the state and reduces the variance of gradient. The policy head outputs the action policy $\pi(\bm{a}^{(t)}|\bm{s}^{(t)})$, which is the probabilities of taking each action $\bm{a}^{(t)}$. The gradient for $\theta_p$ is computed by:
\begin{equation}
d\theta_p \ = \ -\nabla_{\theta_p}\pi(\bm{a}^{(t)}|\bm{s}^{(t)})A(\bm{a}^{(t)}, \bm{s}^{(t)}).
\end{equation}
The two heads are jointly trained in an end-to-end manner.


\section{Experiments}

\subsection{Datasets}
In our paper, we do experiments on three 3D MRI datasets. Each image is cropped based on its non-zero region before used. For each dataset, we access all the image cases having ground truth and split them into several sets. The initial method is defined as the segmentation method of producing the initial coarse segmentations. If we use the images trained on initial method again in update method, the coarse segmentation probability (initial segmentation probability for update method) will be too perfect to be refined in update method because these images have already seen the ground truth in initial method. Therefore, we propose a new way for the dataset splitting: the dataset is split into three parts, two training sets with an equal amount of images and one testing set. In detail, we randomly selected $N_{train}$ cases as the training set for initial method forming $D_{train1}$ and randomly selected another $N_{train}$ cases in the remaining dataset as the training set for update method forming $D_{train2}$. The remaining $N_{test}$ cases are used as testing forming $D_{test}$. Note that the initial segmentation probabilities data for $D_{train2}$ in update method are obtained by testing $D_{train2}$ with initial method. The three datasets are as follows:\\

\noindent \textbf{BraTS2015.}
Brain Tumor Segmentation Challenge 2015 (BraTS) \cite{menze:hal-00935640} provides a dataset for brain tumor segmentation in magnetic resonance images. We use Fluid-attenuated Inversion Recovery (FLAIR) images which contain 274 cases and only segment the whole brain tumor. We set $N_{train}$ as 117 and $N_{test}$ as 40.\\

\noindent \textbf{MM-WHS.}
Multi-Modality Whole Heart Segmentation (MM-WHS) \cite{zhuang2016multi} contains multi-modality whole heart images covering the whole heart substructures. We use the 20 MRI cases and segment the left atrium blood cavity. We set $N_{train}$ as 8 and $N_{test}$ as 4.\\

\noindent \textbf{NCI-ISBI2013.}
NCI-ISBI 2013 Challenge \cite{bloch2015nci} aims at automated segmentation of prostate structures. It provides 80 prostate gland MRI data. We set $N_{train}$ as 32 and $N_{test}$ as 16.

\subsection{Settings}

\noindent \textbf{Evaluation metrics.}
Normally, medical image segmentation is evaluated by the dice score:
\begin{equation}
Dice(S_p, S_g) \ = \ \frac{2|S_p \cap S_g|}{|S_p| + |S_g|},
\end{equation}
where $S_p$ represents the predicted segmentation and $S_g$ represents the ground truth. $|\cdot|$ is the number of voxels in the area.

As we study the interactive image segmentation task, we consider not only the dice score but also the user click number. Our goal is to get a high dice score with a small number of user clicks.\\

\noindent \textbf{User simulation.}
Since it would require large human resources to conduct the experiments with real physicians, we simulate user clicks like other works. While previous works usually give many clicks ($\approx 40$) for training but a few clicks for testing, our interaction policies for training and testing are consistent. Hence, the training setting is similar to that of testing in order to reduce the bias between training and testing. In one training/testing sequence of an image, we give $N_{click}$ clicks each step. Specifically, the clicks are selected as the centers of the largest $N_{click}$ error regions. In addition, a small disturbance $\epsilon_{noise}$ is added to each click point to force the model to be robust and also make it imitate the behavior of a real user.\\

\noindent \textbf{Implementation details.}
For the preprocessing, all the images are normalized by the mean and the standard variation of the whole dataset $D = [D_{train1}, D_{train2}, D_{test}]$. Each image is cropped by the bounding box based on its non-zero region with an extension of $[0, 10]$ voxels and then resized to the size of $55 \times 55 \times 30$. Data augmentation involves flipping in three directions and random rotation with angle range $[-\pi/8, \pi/8]$ in three directions.

As the proposed IteR-MRL can be easily adapted to the interaction-free mode, we firstly train a pure segmentation model IteR-MRL0 as the pretrained model for IteR-MRL. IteR-MRL0 is trained for 1000 epochs and IteR-MRL fine-tunes on IteR-MRL0 for 500 epochs. The learning rate adopts the step decay schedule with an initial learning rate $10^{-4}$. Parameter setting is as follows: $T=5$, $N_{click}=5$, $\gamma=0.95$, $\epsilon_{noise}=[-3,3]^3$. We use Adam algorithm for optimization with minibatch size 1.

The model training time with one Nvidia Titan X GPU varies from several hours to two days for different datasets. The average inference time for each update step is 894ms, which includes 424ms of the interaction simulation time.

\subsection{Results}

For fair comparison, we apply denseCRF to all the models compatible with CRF as the final refinement processing.\\

\noindent \textbf{Comparisons with state-of-the-art methods.}
We compare IteR-MRL with three state-of-the-art methods: Min-Cut \cite{krahenbuhl2011efficient}, DeepIGeoS(R-Net) \cite{wang2018deepigeos} and InterCNN \cite{bredell2018iterative}.

\begin{table}
\centering
\scalebox{0.7}{\begin{tabular}{c | c c c c c} 
\toprule[1pt]
\diagbox{\textbf{Update}}{\textbf{Initial}} & BG & V-Net & HighRes3DNet & DeepIGeoS(P-Net) \\ [0.5ex]
\hline
Initial & 0 & 77.15 & 75.39 & 82.16\\
\hline
Min-cut & 27.46 & 80.69 & 77.05 & 84.08 \\
DeepIGeoS(R-Net) & 82.97 & 85.80 & 85.72 & 84.83 \\ 
InterCNN & 85.17 & 85.56 & 87.29 & 86.54 \\ 
IteR-MRL & \textbf{86.14} & \textbf{88.53} & \textbf{87.43} & \textbf{87.50} \\ [0.5ex] 
\bottomrule[1pt]
\end{tabular}}
\vspace{5pt}
\caption{Combination with different initial methods}
\vspace{2pt}
\label{tab:initial}
\end{table}

\begin{table}
\centering
\scalebox{0.7}{
\begin{tabular}{c c c c c c c} 
\toprule[1pt]
Step & 0 & 1 & 2 & 3 & 4 & 5 \\
Clicks & 0 & 5 & 10 & 15 & 20 & 25 \\ [0.5ex] 
\hline
\multirow{2}*{Min-Cut} & 77.15 & 79.52 & 79.97 & 80.22 & 80.46 & 80.69 \\
 & & (+2.37) & (+0.45) & (+0.25) & (+0.24) & (+0.23) \\
\hline
\multirow{2}*{DeepIGeoS(R-Net)} & 77.15 & \textbf{85.62} & 85.74 & 85.73 & 85.75 & 85.80 \\
 & & (\textbf{+8.47}) & (+0.12) & (-0.01) & (+0.02) & (+0.05) \\
\hline
\multirow{2}*{InterCNN} & 77.15 & 83.19 & 84.39 & 85.16 & 85.52 & 85.56 \\
 & & (+6.04) & (+1.20) & (+0.77) & (+0.36) & (+0.04) \\
\hline
\multirow{2}*{IteR-MRL} & 77.15 & 84.35 & \textbf{86.78} & \textbf{87.61} & \textbf{88.18} & \textbf{88.53} \\ 
 & & (+7.20) & (\textbf{+2.43}) & (\textbf{+0.83}) & (\textbf{+0.57}) & (\textbf{+0.35}) \\[0.5ex] 
\bottomrule[1pt]
\end{tabular}}
\vspace{3pt}
\caption{Performance improvement in one interactive sequence}
\label{tab:iteration}
\end{table}

\begin{figure}
\centering
\includegraphics[width=3.3in]{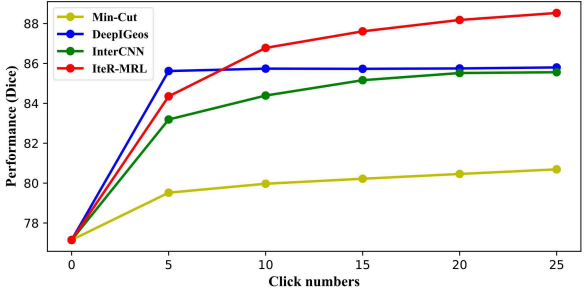}
\caption{Performance improvement shown by curves}
\label{plot}
\end{figure}

In Table \ref{tab:initial}, the update methods receive the coarse segmentations from four different initial segmentation methods: BG (set all voxel labels to background), V-Net \cite{milletari2016v}, HighRes3DNet \cite{li2017compactness} and DeepIGeoS(P-Net) \cite{wang2018deepigeos}. 
The experimental results show that IteR-MRL achieves better performances than the other three methods under each initial method, which shows the robustness and generalization of our method.

\begin{figure}
\centering
\includegraphics[width=2.8in]{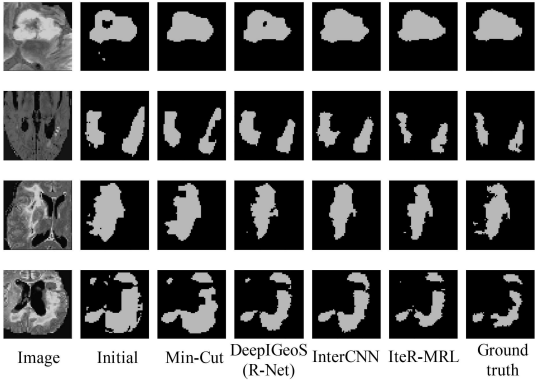}
\caption{Visualization of different update methods}
\label{update-compare}
\end{figure}

\begin{figure*}
\centering
\includegraphics[width=0.9\textwidth]{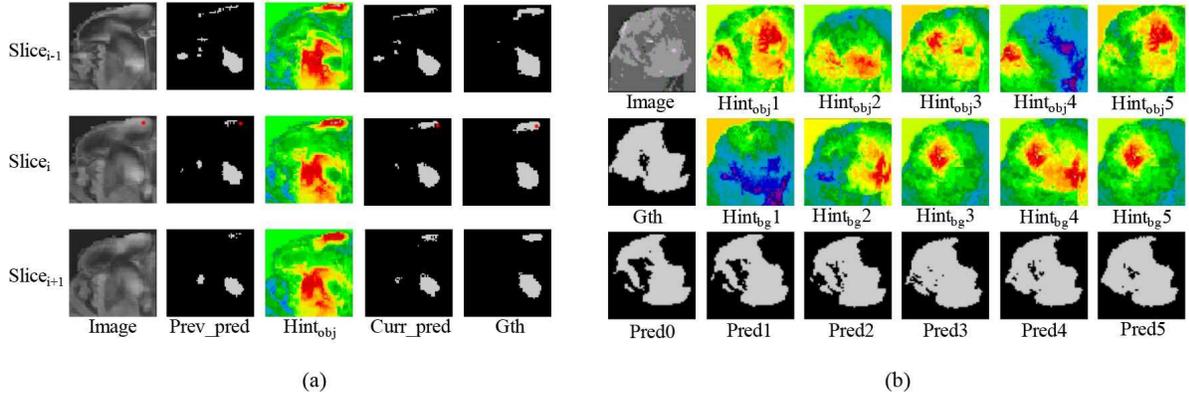}
\caption{The visualization for the relation between predictions and hints. (a) The visualization of one click and its influence on prediction and hint maps. The slice with click and its two neighbor slices are shown. The user click is represented as red points. One row of five figures form a group, which corresponds to one slice [Image, Previous prediction, Object hint map, Current prediction, Ground truth]. (b) The visualization of prediction and hint map for each step. The figures in the first column are [Image, Ground truth, Initial prediction]. Afterwards, each column forms one step, which corresponds to [Object hint map, Background hint map, Prediction].}
\label{ud-ablation}
\end{figure*}

To validate whether considering the relative gain between successive predictions can result in rapid improvements, we also analyze the performance improvement during one refinement sequence in Table \ref{tab:iteration}. We use V-Net here as the initial method (77.15). For the first refinement step, all the update methods have significant improvements in performance (from $+2.37$ to $+8.47$). Starting from the second step, most performances have encountered stagnation (very little improvement) though with newly-added user hints. DeepIGeoS(R-Net) even has a degradation ($-0.01$) at the third step. While the other methods improve slowly at each refinement step, IteR-MRL has a relative high improvement, which proves the effectiveness of considering the relation gain between successive predictions. The large improvement at each refinement step also leads to a good result ($88.53$) in the end. In addition, we notice that IteR-MRL's performance at the second step has already surpassed the others' final performances, achieving a reduction of user click number. Fig. \ref{plot} provides a global view of performance improvement in one interactive sequence.

Fig. \ref{update-compare} gives the visualization of different update methods using V-Net as the initial method. Specifically, we visualize the refined segmentations after five refinement steps. It can be observed that while the other methods tend to produce a rather smooth boundary, IteR-MRL performs better in capturing edge details.

\begin{table}
\centering
\scalebox{0.8}{\begin{tabular}{c c c c}
\toprule[1pt]
Dataset & BRATS2015 & MM-WHS & NCI-ICBI2013 \\
[0.5ex] 
\hline
Initial & 77.15 & 79.60 & 79.34 \\
\hline
Min-Cut & 80.69 & 83.21 & 79.92 \\ 
DeepIGeoS(R-Net) & 85.80 & 85.21 & 79.97 \\ 
InterCNN & 85.56 & 84.76 & 82.14 \\ 
IteR-MRL & \textbf{88.53} & \textbf{86.92} & \textbf{82.71} \\ [0.5ex] 
\bottomrule[1pt]
\end{tabular}}
\vspace{5pt}
\caption{Performances on different datasets}
\vspace{2pt}
\label{tab:data}
\end{table}

The above results are obtained from the experiments with the dataset BraTS2015. More experiments are also conducted on the other two datasets MM-WHS and NCI-ISBI2013 in Table \ref{tab:data} to verify the robustness, with the initial method V-Net. The results prove that IteR-MRL has stable performances on various types of datasets.\\

\begin{table}
\centering
\scalebox{0.8}{\begin{tabular}{c c c c | c c | c} 
\toprule[1pt]
\multicolumn{4}{c|}{Actions} & \multicolumn{2}{c|}{States} & \multirow{2}*{IteR-MRL} \\
\cline{1-6}
$\pm0.1$ & $\pm0.2$ & $\pm0.4$ & $\pm1.0$ & Probability & Binary & \\[0.5ex] 
\hline
& & & \checkmark & & \checkmark & 84.03 \\
& & \checkmark & & \checkmark & & 84.29\\
& \checkmark & & & \checkmark & & 86.51 \\
\checkmark & & & & \checkmark & & 87.20 \\
\checkmark & \checkmark & & & \checkmark & & 87.88 \\
\checkmark & \checkmark & \checkmark & & \checkmark & & \textbf{88.53} \\ 
\checkmark & \checkmark & \checkmark & \checkmark & \checkmark & & 88.02 \\[0.5ex] 
\bottomrule[1pt]
\end{tabular}}
\vspace{5pt}
\caption{Combination of different action and state settings}
\label{tab:settings}
\end{table}

\noindent \textbf{Ablation study.}
We analyze the effect of different action sets to the algorithm performance in Table \ref{tab:settings}. Specially, when the action set only contains ${\pm1.0}$ (line 1), the segmentation probability becomes binary, because the segmentation probability can only take the values 0 and 1. The rest action sets are all designed for the states containing segmentation probability. The influence of the action value and the action number are both analyzed. For the influence of value, we fix the number of actions and let action values vary: we try $\pm1.0$, $\pm0.4$, $\pm0.2$ and $\pm0.1$ (line 1, 2, 3, 4). Comparing the states adopting segmentation probability to those of binary prediction, it can be found that binary prediction has a poor performance caused by the loss of prediction uncertainty. In addition, the results show that small action values have better performances than the larger ones. The reason is that a small action value allows IteR-MRL to make more detailed adjustments, but a large adjustment may over-behave and never reach some specific states. For the influence of the action number, we gradually add new actions to the action set (line 4, 5, 6, 7). It can be observed that abundant actions lead to a better performance by providing IteR-MRL with various degrees of adjustment. In the case with a high confidence, IteR-MRL tends to take a large adjustment, which speeds up the refinement convergence. However, the addition of $\pm1.0$ relatively damages the performance because an adjustment of $\pm1.0$ is too extreme for most of the cases. In general, we learn that the combination of small and large actions except for $\pm1.0$ works best and finally choose $\mathcal{A}=\{\pm0.1, \pm0.2, \pm0.4\}$ as the ideal action set in our model. Note that we have also thought about a continuous action space, but the experimental result shows that it is difficult to train and converge in a continuous action space. In addition, since the final prediction presented to the user is 0 or 1 for each voxel, we are not concerned about the inability to reach the optimal final state with a discrete action space.

\begin{table}
\centering
\scalebox{0.7}{\begin{tabular}{c c c c c c c} 
\toprule[1pt]
Step & 0 & 1 & 2 & 3 & 4 & 5 \\
Clicks & 0 & 5 & 10 & 15 & 20 & 25 \\ [0.5ex] 
\hline
\multirow{2}*{Good Interaction} & 77.15 & \textbf{84.35} & \textbf{86.78} & \textbf{87.61} & \textbf{88.18} & \textbf{88.53} \\ 
 & & (\textbf{+7.20}) & (\textbf{+2.43}) & (\textbf{+0.83}) & (\textbf{+0.57}) & (\textbf{+0.35}) \\
\hline
\multirow{2}*{W/O Interaction} & 77.15 & 78.60 & 79.53 & 80.15 & 80.56 & 80.78 \\ 
 & & (+1.45) & (+0.93) & (+0.62) & (+0.41) & (+0.22) \\
\hline
\multirow{2}*{Bad Interaction} & 77.15 & 76.86 & 75.47 & 74.84 & 74.29 & 72.76 \\ 
 & & (-0.29) & (-1.39) & (-0.63) & (-0.55) & (-1.53) \\[0.5ex] 
\bottomrule[1pt]
\end{tabular}}
\vspace{5pt}
\caption{Contribution of interactions to performance}
\label{tab:interaction}
\end{table}

As we know, the interaction and the model can both lead to the improvement of performance. We now analyze how much the interaction contributes to the performance by changing the interaction strategy. In addition to the good interaction used before, two more comparative experiments are done in Table \ref{tab:interaction}. The one without interaction is to always fill the hint maps with random noise and the model will not receive any new interactive information. The other one with bad interaction is to randomly choose the user click points among all the voxels. In this case, the interaction may pass the wrong message to the model. As a result, we find that the meaningful interaction does help greatly improve the performance. It can also be observed that the one without interactions still has some gain of performance, which may come from the iterative model itself. Moreover, the degradation of the one with bad interactions indicates that ineffective interaction can damage the performance.

Fig. \ref{ud-ablation} presents the visualization for the relation between predictions and hints. Fig. \ref{ud-ablation}(a) shows the influence of user interaction on prediction and hint maps. Since the data is 3D, we show the slice with click (the middle row) and its two neighbor slices (rows on both sides). The red parts on hint maps are the recommended object regions. We find that the proposed algorithm can successfully correct the local region around the user click (the red point). Besides, the corresponding regions on neighbor slices are also improved. In Fig. \ref{ud-ablation}(b), we observe the change of predictions and hint maps in one interactive sequence which contains five steps. The user clicks are not shown because the slices with clicks vary at each step and we only focus on changes of one specific slice. With the indications of hint maps, IteR-MRL succeeds in refining the initial prediction step by step.

\section{Conclusion}
In this paper, we propose a novel iteratively-refined interactive segmentation method for 3D medical images using multi-agent reinforcement learning. 
The method explicitly models the dynamic process of interactive image segmentation task in order to get a rapid segmentation improvement at each iteration. 
In addition, it augments the state space of MDP with the segmentation probability, allowing the agent to make finer adjustments than the conventional binary approach. 
The experimental results show that it performs better than the state-of-the-art methods and it is robust to different initial segmentations and various datasets.

{\small
\bibliographystyle{ieee_fullname}
\bibliography{egbib}

\begin{thebibliography}{10}\itemsep=-1pt

\bibitem{acuna2018efficient}
David Acuna, Huan Ling, Amlan Kar, and Sanja Fidler.
\newblock Efficient interactive annotation of segmentation datasets with
  {P}olygon-{RNN}++.
\newblock In {\em CVPR}, pages 859--868, 2018.

\bibitem{bloch2015nci}
N Bloch, A Madabhushi, H Huisman, J Freymann, J Kirby, M Grauer, A Enquobahrie,
  C Jaffe, L Clarke, and K Farahani.
\newblock {NCI-ISBI} 2013 challenge: automated segmentation of prostate
  structures.
\newblock {\em The Cancer Imaging Archive}, 370, 2015.

\bibitem{boykov2004experimental}
Yuri Boykov and Vladimir Kolmogorov.
\newblock An experimental comparison of {M}in-{C}ut/{M}ax-{F}low algorithms for
  energy minimization in vision.
\newblock {\em IEEE Transactions on Pattern Analysis \& Machine Intelligence},
  (9):1124--1137, 2004.

\bibitem{boykov2001interactive}
Yuri~Y Boykov and M-P Jolly.
\newblock Interactive graph cuts for optimal boundary \& region segmentation of
  objects in nd images.
\newblock In {\em ICCV}, pages 105--112, 2001.

\bibitem{bredell2018iterative}
Gustav Bredell, Christine Tanner, and Ender Konukoglu.
\newblock Iterative interaction training for segmentation editing networks.
\newblock In {\em MLMI International Workshop}, pages 363--370, 2018.

\bibitem{castrejon2017annotating}
Lluis Castrejon, Kaustav Kundu, Raquel Urtasun, and Sanja Fidler.
\newblock Annotating object instances with a {P}olygon-{RNN}.
\newblock In {\em CVPR}, pages 5230--5238, 2017.

\bibitem{cciccek20163d}
{\"O}zg{\"u}n {\c{C}}i{\c{c}}ek, Ahmed Abdulkadir, Soeren~S Lienkamp, Thomas
  Brox, and Olaf Ronneberger.
\newblock {3D U-Net}: learning dense volumetric segmentation from sparse
  annotation.
\newblock In {\em MICCAI}, pages 424--432, 2016.

\bibitem{criminisi2008geos}
Antonio Criminisi, Toby Sharp, and Andrew Blake.
\newblock Geos: Geodesic image segmentation.
\newblock In {\em ECCV}, pages 99--112, 2008.

\bibitem{furuta2019fully}
Ryosuke Furuta, Naoto Inoue, and Toshihiko Yamasaki.
\newblock Fully convolutional network with multi-step reinforcement learning
  for image processing.
\newblock In {\em AAAI}, pages 3598--3605, 2019.

\bibitem{krahenbuhl2011efficient}
Philipp Kr{\"a}henb{\"u}hl and Vladlen Koltun.
\newblock Efficient inference in fully connected {CRF}s with gaussian edge
  potentials.
\newblock In {\em NeurIPS}, pages 109--117, 2011.

\bibitem{li2017compactness}
Wenqi Li, Guotai Wang, Lucas Fidon, Sebastien Ourselin, M~Jorge Cardoso, and
  Tom Vercauteren.
\newblock On the compactness, efficiency, and representation of {3D}
  convolutional networks: brain parcellation as a pretext task.
\newblock In {\em IPMI}, pages 348--360, 2017.

\bibitem{menze:hal-00935640}
Bjoern Menze, Andras Jakab, Stefan Bauer, Jayashree Kalpathy-Cramer, Keyvan
  Farahani, Justin Kirby, Yuliya Burren, Nicole Porz, Johannes Slotboom, Roland
  Wiest, Levente Lanczi, Elisabeth Gerstner, et~al.
\newblock The multimodal brain tumor image segmentation benchmark ({BraTS}).
\newblock {\em IEEE Transactions on Medical Imaging}, page~33, 2014.

\bibitem{milletari2016v}
Fausto Milletari, Nassir Navab, and Seyed-Ahmad Ahmadi.
\newblock {V-Net}: Fully convolutional neural networks for volumetric medical
  image segmentation.
\newblock In {\em 3DV}, pages 565--571, 2016.

\bibitem{mnih2016asynchronous}
Volodymyr Mnih, Adria~Puigdomenech Badia, Mehdi Mirza, Alex Graves, Timothy
  Lillicrap, Tim Harley, David Silver, and Koray Kavukcuoglu.
\newblock Asynchronous methods for deep reinforcement learning.
\newblock In {\em ICML}, pages 1928--1937, 2016.

\bibitem{rajchl2016deepcut}
Martin Rajchl, Matthew~CH Lee, Ozan Oktay, Konstantinos Kamnitsas, Jonathan
  Passerat-Palmbach, Wenjia Bai, Mellisa Damodaram, Mary~A Rutherford, Joseph~V
  Hajnal, Bernhard Kainz, et~al.
\newblock Deepcut: Object segmentation from bounding box annotations using
  convolutional neural networks.
\newblock {\em IEEE Transactions on Medical Imaging}, 36(2):674--683, 2016.

\bibitem{ronneberger2015u}
Olaf Ronneberger, Philipp Fischer, and Thomas Brox.
\newblock {U-Net}: Convolutional networks for biomedical image segmentation.
\newblock In {\em MICCAI}, pages 234--241, 2015.

\bibitem{rother2004grabcut}
Carsten Rother, Vladimir Kolmogorov, and Andrew Blake.
\newblock Grab{C}ut: Interactive foreground extraction using iterated graph
  cuts.
\newblock In {\em ACM Transactions on Graphics}, volume~23, pages 309--314,
  2004.

\bibitem{song2018seednet}
Gwangmo Song, Heesoo Myeong, and Kyoung Mu~Lee.
\newblock Seednet: Automatic seed generation with deep reinforcement learning
  for robust interactive segmentation.
\newblock In {\em CVPR}, pages 1760--1768, 2018.

\bibitem{wang2018interactive}
Guotai Wang, Wenqi Li, Maria~A Zuluaga, Rosalind Pratt, Premal~A Patel, Michael
  Aertsen, Tom Doel, Anna~L David, Jan Deprest, S{\'e}bastien Ourselin, et~al.
\newblock Interactive medical image segmentation using deep learning with
  image-specific fine tuning.
\newblock {\em IEEE Transactions on Medical Imaging}, 37(7):1562--1573, 2018.

\bibitem{wang2018deepigeos}
Guotai Wang, Maria~A Zuluaga, Wenqi Li, Rosalind Pratt, Premal~A Patel, Michael
  Aertsen, Tom Doel, Anna~L David, Jan Deprest, S{\'e}bastien Ourselin, et~al.
\newblock {DeepIGeoS}: a deep interactive geodesic framework for medical image
  segmentation.
\newblock {\em IEEE Transactions on Pattern Analysis and Machine Intelligence},
  41(7):1559--1572, 2018.

\bibitem{wang2016slic}
Guotai Wang, Maria~A Zuluaga, Rosalind Pratt, Michael Aertsen, Tom Doel, Maria
  Klusmann, Anna~L David, Jan Deprest, Tom Vercauteren, and S{\'e}bastien
  Ourselin.
\newblock Slic-seg: A minimally interactive segmentation of the placenta from
  sparse and motion-corrupted fetal {MRI} in multiple views.
\newblock {\em Medical Image Analysis}, 34:137--147, 2016.

\bibitem{xu2016deep}
Ning Xu, Brian Price, Scott Cohen, Jimei Yang, and Thomas~S Huang.
\newblock Deep interactive object selection.
\newblock In {\em CVPR}, pages 373--381, 2016.

\bibitem{zhou2018unet++}
Zongwei Zhou, Md~Mahfuzur~Rahman Siddiquee, Nima Tajbakhsh, and Jianming Liang.
\newblock {UNet}++: A nested {U-Net} architecture for medical image
  segmentation.
\newblock In {\em DLMIA}, pages 3--11. 2018.

\bibitem{zhuang2016multi}
Xiahai Zhuang and Juan Shen.
\newblock Multi-scale patch and multi-modality atlases for whole heart
  segmentation of {MRI}.
\newblock {\em Medical Image Analysis}, 31:77--87, 2016.

\end{thebibliography}
}

\end{document}